\documentclass[sigconf]{aamas} 

\usepackage{balance} 
\usepackage{graphicx}
\usepackage{algpseudocode,algorithm}
\usepackage{algorithmicx}
\algrenewcommand\textproc{}
\usepackage[font={small}]{caption, subfig}
\usepackage{microtype}
\usepackage{multicol}
\usepackage{amsthm} 
\usepackage{thmtools}
\usepackage{paralist}
\usepackage{enumitem}
\usepackage{autoaligne}
\usepackage{multirow}
\usepackage{xspace}
\usepackage{pgf, tikz}
\usetikzlibrary{arrows, automata}

\algdef{SE}[DOWHILE]{Do}{doWhile}{\algorithmicdo}[1]{\algorithmicwhile\ #1}%

\setcopyright{none}
\renewcommand\footnotetextcopyrightpermission[1]{}
\title{Argument Schemes and Dialogue for Explainable Planning}

\author{Quratul-ain Mahesar}
\affiliation{
  \institution{Department of Informatics, King's College London, London, UK}}
\email{quratul-ain.mahesar@kcl.ac.uk}

\author{Simon Parsons}
\affiliation{
  \institution{School of Computing Science, University of Lincoln, Lincoln, UK}}
\email{sparsons@lincoln.ac.uk}

\begin{abstract}
Artificial Intelligence (AI) is being increasingly deployed in practical applications. However, there is a major concern whether AI systems will be trusted by humans. In order to establish trust in AI systems, there is a need for users to understand the reasoning behind their solutions. Therefore, systems should be able to explain and justify their output. In this paper, we propose an argument scheme-based approach to provide explanations in the domain of AI planning. We present novel argument schemes to create arguments that explain a plan and its key elements; and a set of critical questions that allow interaction between the arguments and enable the user to obtain further information regarding the key elements of the plan. Furthermore, we present a novel dialogue system using the argument schemes and critical questions for providing interactive dialectical explanations. 
\end{abstract}


\newcommand{\BibTeX}{\rm B\kern-.05em{\sc i\kern-.025em b}\kern-.08em\TeX}

\begin{document}


\pagestyle{fancy}
\fancyhead{}


\maketitle 


\section{Introduction}
Artiﬁcial intelligence (AI) researchers are increasingly concerned that whether the systems they build will be trusted by humans. One mechanism for increasing trust is to make AI systems capable of explaining their reasoning. In this paper we provide a mechanism for explaining the output of an AI planning system. Automated planning~\cite{daglib04} is one of the sub fields of AI that focuses on developing techniques to create efficient plans, i.e., sequences of actions that should be performed in order to achieve a set of goals. In practical applications, for instance, this set of actions can be passed to a robot, or a manufacturing system, that can follow the plan and produce the desired result. 

Explainable AI Planning (XAIP)~\cite{fox2017explainable} is a field that involves explaining AI planning systems to a user. The main goal of a plan explanation is to help humans understand reasoning behind the plans produced by the planners. Approaches to this problem include explaining planner decision-making processes as well as forming explanations from the models. Previous work on model-based explanations of plans include~\cite{fox2017explainable,smith12}.

To provide explanations for plans, we make use of argumentation. Argumentation~\cite{SR2009} is a logical model of reasoning that is aimed at the evaluation of possible conclusions or claims by considering reasons for and against them. These reasons, i.e., arguments and counter-arguments, provide support for and against the conclusions or claims, through a combination of dialectical and logical reasoning. Argumentation is connected to the idea of establishing trust in AI systems by explaining the results and processes of the computation of a solution or decision, and has been used in many applications in multi-agent planning~\cite{Torre_o_2018} and practical reasoning~\cite{AtkinsonB07}. Argumentation has also been used in explanation dialogues. A dialogue system~\cite{BexW16} for argumentation and explanation consists of a communication language that defines the speech acts and protocols that allow transitions in the dialogue. This allows the explainee to challenge and interrogate the given explanations to gain further understanding.

In this paper, we show how to use argumentation to generate explanations in the domain of AI planning, to answer questions such as ‘Why a?’, where a is an action in the plan, or ‘How g?’, where g is a goal. Questions like these are inherently based upon definitions held in the domain related to a particular problem and solution. Furthermore, questions regarding particular state information may arise, such as ‘Why a here?’. To answer such questions, it is necessary to extract relevant information about actions, states and goals from the planning model. This information allows us to provide supporting evidence to draw conclusions within the explanation process. In addition, it allows us to create a summary explanation of the whole plan and, through dialogue and questioning, extract further information regarding the elements of the plan.

Our approach is built around a set of argument schemes~\cite{walton1996argumentation} which create arguments that explain and justify a plan and its key elements (i.e., actions, states and goals). 
We call such arguments \emph{plan explanation arguments}.
These schemes are complemented by critical questions that allow a user to seek further information regarding the plan elements, and allow interaction between different arguments. 
Plan explanation arguments can be constructed through the instantiation of argument schemes, and can be questioned with a given scheme’s associated critical questions (CQs). 
Given a plan explanation argument $A$ instantiating an argument scheme, CQs are possible counter-arguments to $A$, and they question the premises, i.e., presumptions of $A$ about the key elements of the plan, and so shift the burden of proof such that further arguments must be put forward to argue for the plan elements in the premises questioned. 
Until this burden of proof is met, a plan explanation argument $A$ cannot be said to be justified. 
The plan explanation arguments enable a planning system to answer such questions at different levels of granularity. 
In addition, we present a dialogue system utilizing the argument schemes and critical questions for providing an interactive approach for explanation and query answering, and give algorithms that describe the mechanization of a dialogue conversation between the planner and user in the dialogue system.
To make our argumentation-based explanations for the planning study concrete, we use a version of the classic blocks world. 

\section{Related Work}
\label{sec:related-work}

Our research is inspired by work in practical reasoning and argumentation for multi-agent planning. However, our argument scheme based approach, generates explanations for a plan created by an AI planner, which we assume to be a single entity. One of the most well known scheme-based approaches in practical reasoning is presented in~\cite{AtkinsonB07}, which is accompanied by a set of critical questions that allow agents to evaluate the outcomes on the basis of the social values highlighted by the arguments. \cite{Medellin-GasqueABM13} is an extension of~\cite{AtkinsonB07} specifically designed for multi-agent planning where agents can discuss the suitability of plans based on an argumentation scheme and associated critical questions. Furthermore, in~\cite{TonioloNS11}, a model for arguments is presented that contributes in deliberative dialogues based on argumentation schemes for arguing about norms and actions in a multi-agent system. \cite{Oren13} has proposed a similar scheme-based approach for normative practical reasoning where arguments are constructed for a sequence of actions. 

\cite{ShamsVOP20} propose a framework that integrates both the reasoning and dialectical aspects of argumentation to perform normative practical reasoning, enabling an agent to act in a normative environment under conflicting goals and norms and generate explanation for agent behaviour. 
\cite{BelesiotisRR10} explore the use of situation calculus as a language to present arguments about a common plan in a multi-agent system and \cite{TangP05} present an argumentation-based  approach  to deliberation, the process by which two or more agents reach a consensus on a course of action. \cite{pardo2011multiagent} propose a formal model of argumentative dialogues for multi-agent planning, with a focus on cooperative planning, and \cite{ferrando2012defeasible} present a practical solution for multi-agent planning based upon an argumentation-based defeasible planning framework on ambient intelligence applications.

The works that are closest to our research for generating plan explanations using argumentation are given in~\cite{Caminada2014ScrutablePE,OrenDV20} and~\cite{fan18}. In~\cite{Caminada2014ScrutablePE,OrenDV20}, a dialectical proof based on the grounded semantics~\cite{CaminadaP12} is created to justify the actions executed in a plan. More recently, in~\cite{fan18}, an assumption-based argumentation framework (ABA)~\cite{aba_Dung2009} is used to model the planning problem and generate explanation using the related admissible semantic~\cite{aaai-FanT15}. Our work differs from both, since we present argument schemes to generate the explanation arguments for all the key elements of the plan, and critical questions to allow interaction between the arguments. Whilst previous research provides a static explanation, in our approach, a dialogue system is presented that allows the user to engage in a dialogue conversation with the AI planner to challenge and interrogate the planner explanations.

\section{Argumentation Model}
\label{sec:argumentation-model}

In this section, we describe plan explanation arguments and their interactions at the level of abstract argumentation~\cite{Dung95onthe}, together with a set of critical questions that are suitable for arguing over plan explanations as we will later show.

An \textit{argumentation framework} is simply a set of arguments and a binary attack relation among them. Given an argumentation framework, argumentation theory allows to identify the sets of arguments that can survive the conflicts expressed in the framework. 

\begin{definition}{(Abstract Argumentation Framework~\cite{Dung95onthe}):}
	\label{def:arg-framework}
	An abstract argumentation framework (AAF) is a pair $\mathit{AAF} = (\mathcal{A}, \mathcal{R})$, where $\mathcal{A}$ is a set of arguments and $\mathcal{R}$ is an attack relation $(\mathcal{R} \subseteq \mathcal{A} \times \mathcal{A})$. The notation $(A,B) \in \mathcal{R}$ where $A,B \in \mathcal{A}$ denotes that \textit{argument A attacks argument B}.
\end{definition}

Dung~\cite{Dung95onthe} originally introduced an extension approach to define the acceptability of arguments, i.e., semantics for an abstract argumentation framework. An extension is a subset of $\mathcal{A}$ that represents the set of arguments that can be accepted together. For an $\mathit{AAF} = (\mathcal{A}, \mathcal{R})$:
\begin{inparaenum}
\item A set $\mathcal{E} \subseteq \mathcal{A}$ is said to be \textit{conflict free} if and only if there are no $A, B \in \mathcal{E}$ such that $(A,B) \in \mathcal{R}$. 
\item A set $\mathcal{E} \subseteq \mathcal{A}$ is said to be \textit{admissible} if and only if it is conflict free and defends all its arguments. $\mathcal{E}$ defends $A$ if and only if for every argument $B \in \mathcal{A}$, if we have $(B,A) \in \mathcal{R}$ then there exists $C \in \mathcal{E}$ such that $(C,B) \in \mathcal{R}$. 
\item A set $\mathcal{E} \subseteq \mathcal{A} $ is a \textit{complete} extension if and only if $\mathcal{E}$ is an admissible set which contains all the arguments it defends.
\item A set $\mathcal{E} \subseteq \mathcal{A} $ is a \textit{grounded} extension if and only if $\mathcal{E}$ is a minimal (for set inclusion) complete extension.
\end{inparaenum}
Below, we will show how to formulate the explanation of a plan (and its key elements) as an argument in such a way that the argument is only acceptable if the plan is valid.
We then adopt the grounded semantics to establish acceptability so that the explanation argument will only be acceptable if none of the objections, established using critical questions, are supported by the planning model.

\section{Planning Model}
\label{sec:planning-background}

In this section, we introduce the planning model that we use. This is based on an instance of the most widely used planning representation, PDDL~\cite{book-pddl}.

The main components are:
\begin{definition}{(Planning Problem)}
A planning problem is a tuple $P = \langle O, \mathit{Pr}, \bigtriangleup_I, \bigtriangleup_G, A, \Sigma, G \rangle$, where:
\begin{enumerate}
    \item $O$ is a set of objects;
    \item $\mathit{Pr}$ is a set of predicates;
    \item $\bigtriangleup_I \subseteq \mathit{Pr}$ is the initial state;
    \item $\bigtriangleup_G \subseteq \mathit{Pr}$ is the goal state, and $G$ is the set of goals;   
    \item $A$ is a finite, non-empty set of actions;
    \item $\Sigma$ is the state transition system;    
\end{enumerate}
\end{definition}

\begin{definition}{(Predicates)}
$\mathit{Pr}$ is a set of domain predicates, i.e., properties of objects that we are interested in, that can be true or false. For a state $s \subseteq Pr$, $s^+$ are predicates considered \textit{true}, and $s^- = Pr\setminus s^+$. A state $s$ satisfies predicate $pr$, denoted as $s \models pr$, if $pr \in s$, and satisfies predicate $\neg pr$, denoted $s \models \neg pr$, if $pr \not\in s$.
\end{definition}

\begin{definition}{(Action)}
An action $a = \langle pre, post\rangle$ is composed of sets of predicates $pre$, $post$ that represent $a$'s pre and post conditions respectively. Given an action $a=\langle pre, post\rangle$, we write $pre(a)$ and $post(a)$ for $pre$ and $post$.
Postconditions are divided into $add(post(a)^+)$ and $delete(post(a)^-)$ postcondition sets. An action $a$ can be executed in state $s$ iff the state satisfies its preconditions. The postconditions of an action are applied in the state $s$ at which the action ends, by adding $post(a)^+$ and deleting $post(a)^-$.
\end{definition}

\begin{definition}{(State Transition System)} 
The state-transition system is denoted by $\Sigma =(S,A,\gamma)$, where:
\begin{itemize}
    \item $S$ is the set of states.
    \item $A$ is a finite, non-empty set of actions.
    \item $\gamma: S \times A \rightarrow S$ where:
        \begin{itemize}
            \item $\gamma(S,a) \rightarrow (S \setminus post(a)^-)) \cup post(a)^+$, if $a$ is applicable in $S$;
            \item $\gamma(S,a) \rightarrow \mathit{undefined}$ otherwise;
            \item $S$ is closed under $\gamma$.
        \end{itemize}
\end{itemize}
\end{definition}

\begin{definition}{(Goal)}
A goal achieves a certain state of affairs. Each $g \in G$ is a set of predicates $g=\{r_1,...,r_n\}$, known as \textit{goal requirements} (denoted as $r_i$), that should be satisfied in the state to satisfy the goal.
\end{definition}

We then define a plan as follows.

\begin{definition}{(Plan)}
A plan $\pi$ is a sequence of actions $\langle a_1,...,a_n \rangle$. 
A plan $\pi$ is a solution to a planning problem $P$, i.e., plan $\pi$ is \emph{valid} iff:
\begin{inparaenum}
    \item Only the predicates in $\bigtriangleup_I$ hold in the initial state: $S_1 = \bigtriangleup_I$;
    \item the preconditions of action $a_i$ hold at state $S_i$, where $i=1,2,...,n$;
    \item $\gamma(S,\pi)$ satisfies the set of goals $G$.
    \item the set of goals satisfied by plan $\pi$ is a non-empty $G_\pi \neq \emptyset$ consistent subset of goals.
\end{inparaenum}
\end{definition}

Finally we define the state transitions associated with a plan.

\begin{definition}{(Extended State Transition System)}
The extended state transition function for a plan is defined as follows:
\begin{itemize}
    \item $\gamma(S, \pi) \rightarrow S$ if $|\pi|=0$ (i.e., if $\pi$ is empty);
    \item $\gamma(S, \pi) \rightarrow \gamma(\gamma(S,a_1),a_2,...,a_n)$ if $|\pi|>0$ and $a_1$ is applicable in $S$;
    \item $\gamma(S, \pi) \rightarrow \mathit{undefined}$ otherwise.
\end{itemize}
\end{definition}
Each action in the plan can be performed in the state that results from the application of the previous action in the sequence. After performing the final action, the set of goals $G_\pi$ will be true.
We present the following Blocks World example to illustrate.

\begin{example}
\label{example:blocks-world}

A classic blocks world consists of the following: A flat surface such as a tabletop; an adequate set of identical blocks which are identified by letters; and the blocks can be stacked one on one to form towers of unlimited height. 
We have three predicates to capture the domain: $\mathit{On(X,Y)}$, block $X$ is on block $Y$; $\mathit{Ontable(X)}$, block $X$ is on the table; and $\mathit{Clear(X)}$, block $X$ has nothing on it. 
We have two actions,  $a_1$ and $a_2$:
\begin{enumerate}
    \item $a_1: \mathit{Unstack(X,Y)}$ -- pick up clear block $X$ from block $Y$; 
        \begin{itemize}
            \item $\mathit{pre(a_1)}: \{ \mathit{Clear(X)}, \mathit{On(X,Y)} \}$
            \item $\mathit{post(a_1)^+}: \{ \mathit{Ontable(X)}, \mathit{Clear(Y)} \}$
            \item $\mathit{post(a_1)^-}: \{ \mathit{On(X,Y)} \}$            
        \end{itemize}
    \item $a_2: \mathit{Stack(X,Y)}$ -- place block $X$ onto clear block $Y$; 
        \begin{itemize}
            \item $\mathit{pre(a_2)}: \{ \mathit{Ontable(X)}, \mathit{Clear(X)}, \mathit{Clear(Y)} \}$
            \item $\mathit{post(a_2)^+}: \{ \mathit{On(X,Y)} \}$
            \item $\mathit{post(a_2)^-}: \{ \mathit{Ontable(X)}, \mathit{Clear(Y)} \}$            
        \end{itemize}    
\end{enumerate}

\noindent The initial and goal states of the blocks world problem are shown in Figure~\ref{fig:blocks_world}.
\begin{figure}[t]
	\centering
\scalebox{0.8}{\begin{tikzpicture}
 
\filldraw[draw=black,fill=darkgray] (0,0) rectangle node [below] {Initial State} (3,0.25);
\draw[black, very thick] (1,0.25) rectangle node{C} (2,1.00);
\draw[black, very thick] (1,1.00) rectangle node{B} (2,1.75);
\draw[black, very thick] (1,1.75) rectangle node{A} (2,2.50);

\draw[->, line width=2mm] (3,1.5) -- (4.5,1.5);

\filldraw[draw=black,fill=darkgray] (4.5,0) rectangle node [below] {Goal State} (8.5,0.25);
\draw[black, very thick] (5,0.25) rectangle node{A} (6,1.00);
\draw[black, very thick] (5,1.00) rectangle node{C} (6,1.75);

\draw[black, very thick] (7,0.25) rectangle node{B} (8,1.00);
\end{tikzpicture}}
	\caption{Blocks World Example}
	\label{fig:blocks_world}
\end{figure}
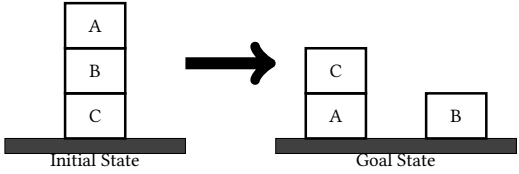

\noindent The initial state $\bigtriangleup_I$ is given by
\[\{ \mathit{Ontable(C)}, \mathit{On(B,C)}, \mathit{On(A,B)},  \allowbreak\mathit{Clear(A)} \}\] 
and the goal state $\bigtriangleup_G$ is given by \[\{ \mathit{On(C,A)}, \mathit{Ontable(A)},\allowbreak \mathit{Ontable(B)}, \allowbreak\mathit{Clear(C)},\allowbreak \mathit{Clear(B)} \}\]
The action sequence: 
$\langle \mathit{Unstack(A,B)}, \mathit{Unstack(B,C)}, \mathit{Stack(C,A)} \rangle$ is a valid plan.
\hfill$\Box$
\end{example}

\section{Argument Schemes for Explaining Plans}
\label{sec:argument-schemes}

In scheme-based approaches~\cite{walton1996argumentation} arguments are expressed in natural language and a set of critical questions is associated with each scheme, identifying how the scheme can be attacked. Below, we introduce a set of argument schemes for explaining a plan and its key elements, i.e., action, state and goal. The set of critical questions allow a user to ask for a summary explanation for the plan and consequently interrogate the elements of the plan by questioning the premises of the arguments put forward by the planner. The explanation arguments constructed using the argument schemes allow the planner to answer any user questions.

\begin{definition}
Given a planning problem $P$:
\begin{itemize}
    \item $\mathit{HoldPrecondition(pre(a),S)}$ denotes that the precondition $pre(a)$ of action $a$ holds at the state $S$.
    \item $\mathit{HoldGoal(g, S)}$ denote that the goal $g$ holds at the state $S$.
    \item $\mathit{HoldGoals(G, \bigtriangleup_G)}$ denotes that all the goals in the set of goals $G$ hold at the goal state $\bigtriangleup_G$.
    \item $\mathit{ExecuteAction(a,S)}$ denotes that action $a$ is executed at state $S$.
    \item $\mathit{AchieveGoal(a,g)}$ denotes that action $a$ achieves goal $g$.
    \item $\mathit{AchieveGoals(\pi,G)}$ denotes that sequence of actions $\pi$ achieves the set of goals $G$.
    \item $\mathit{Solution(\pi, P)}$ denotes that $\pi$ is a solution to the planning problem $P$.
\end{itemize}
\end{definition}

\begin{definition}{(Action Argument Scheme $\mathit{Arg}_a$)}
An action argument $\mathit{Arg}_a$ explains how it is possible to execute an action $a$:
\begin{itemize}
    \item \textbf{Premise $1$:} $\mathit{HoldPrecondition(pre(a), S_1)}$. In the current state $S_1$, the pre-condition $pre(a)$ of action $a$ holds.
    \item  \textbf{Conclusion}: $\mathit{ExecuteAction(a, S_1)}$. Therefore, we can execute action $a$ in the current state $S_1$.    
\end{itemize}
\end{definition}

\begin{example}
\label{example:action-as}

We consider the blocks world of Example~\ref{example:blocks-world}. The explanation argument for the first action $\mathit{Unstack(A,B)}$ is shown as follows. Where:
\begin{itemize}
    \item $pre(\mathit{Unstack(A,B)}) = \{ \mathit{Clear(A), On(A,B)} \}$.
    \item $S_1 = \{ \mathit{Ontable(C), On(B,C), On(A,B), Clear(A)} \}$.
\end{itemize}

     \noindent \textbf{Premise $1$:} 
     
     \noindent $\mathit{HoldPrecondition(pre(\mathit{Unstack(A,B)}), S_1)}$
     
     \smallskip
     \noindent In the current state $\{ \mathit{Ontable(C), On(B,C), On(A,B), Clear(A)} \}$, the pre-condition $\{ \mathit{Clear(A), On(A,B)} \}$ of action $\mathit{Unstack(A,B)}$ holds.

    \smallskip
    \noindent \textbf{Conclusion:} 
    
    \noindent $\mathit{ExcecuteAction(Unstack(A,B), \; S_1)}$

    \smallskip
    \noindent Therefore, we can execute action $\mathit{Unstack(A,B)}$ in the current state $\{ \mathit{Ontable(C), On(B,C), On(A,B), Clear(A)} \}$.    

\hfill$\Box$
\end{example}

\begin{definition}{(State Argument Scheme $\mathit{Arg}_S$)}\footnote{This does not apply to the initial state $\bigtriangleup_I$ and we assume that the user knows the initial state is true by default.}
A state argument $\mathit{Arg}_S$ explains how the state $S$ becomes true:
\begin{itemize}
    \item \textbf{Premise $1$:} 
    $\gamma(S_1,a) \rightarrow ( (S_1 \setminus post(a)^-)) \cup post(a)^+ = S )$.
    In the current state $S_1$, we can execute the action $a \in \pi$, after which the negative postconditions $post(a)^-$ do not hold and the positive postconditions $post(a)^+$ hold, that results in the state $S$.
    \item \textbf{Conclusion:} Therefore, the state $S$ is true.
\end{itemize}
\end{definition}

\begin{example}
\label{example:state-as}

The state argument $\mathit{Arg}_S$ for the state $ S = \{ \mathit{On(B,C),}$

\noindent $\mathit{Clear(A), Clear(B), Ontable(A), Ontable(C)} \}$ in the Example~\ref{example:blocks-world} is shown as follows. Where:

\begin{itemize}
    \item $a = \mathit{Unstack(A,B)}$.
    \item $post(a)^- = \{ \mathit{On(A,B)} \}$
    \item $post(a)^+ = \{ \mathit{Ontable(A), Clear(B)} \}$
    \item $S_1 = \{ \mathit{Ontable(C), On(B,C), On(A,B), Clear(A)} \}$.
\end{itemize}

     \noindent \textbf{Premise $1$:} 
     
     \noindent $\gamma(S_1, \; \mathit{Unstack(A,B)}) \rightarrow ( (S_1 \setminus post(a)^-)) \cup post(a)^+ = S )$.
     
     \smallskip
     \noindent In the current state $\{ \mathit{Ontable(C), On(B,C), On(A,B), Clear(A)} \}$, we can execute the action $\mathit{Unstack(A,B)}$, after which the negative postconditions $\{ \mathit{On(A,B)} \}$ do not hold and the positive postconditions 
     
     \noindent $\{ \mathit{Ontable(A), Clear(B)} \}$ hold, that results in the state $\{ \mathit{On(B,C),}$
     
     \noindent $\mathit{Clear(A), Clear(B), Ontable(A), Ontable(C)} \}$.
     
     \smallskip
     \noindent \textbf{Conclusion:} 
     
     \noindent Therefore, the state $\{ \mathit{On(B,C), Clear(A), Clear(B), Ontable(A),}$
     
     \noindent $\mathit{Ontable(C)} \}$ is true.
\hfill$\Box$
\end{example}

\begin{definition}{(Goal Argument Scheme $\mathit{Arg}_g$)}
A goal argument $\mathit{Arg}_g$ explains how a goal is achieved by an action in the plan:

\begin{itemize}
    \item \textbf{Premise $1$:} $\gamma(S_1,a) \rightarrow S_2$. In the current state $S_1$, we can execute the action $a \in \pi$, that results in the next state $S_2$.
    \item \textbf{Premise $2$:} $\mathit{HoldGoal(g, S_2)}$. In the next state $S_2$, the goal $g$ holds.
    \item \textbf{Conclusion:} $\mathit{AchieveGoal(a,g)}$: Therefore, the action $a$ achi\-e\-ves the goal $g$.    
\end{itemize}
\end{definition}

\begin{example}
\label{example:goal-as}

The goal argument $\mathit{Arg}_g$ for the goal $g = \mathit{Ontable(A)}$ in the Example~\ref{example:blocks-world} is shown as follows. Where:

\begin{itemize}
    \item $a = \mathit{Unstack(A,B)}$.
    \item $S_1 = \{ \mathit{Ontable(C)}, \mathit{On(B,C)}, \mathit{On(A,B)}, \mathit{Clear(A)} \}$.
    \item $S_2 = \{ \mathit{Ontable(C)}, \mathit{On(B,C)}, \mathit{Clear(A)}, \mathit{Clear(B)}, \mathit{Ontable(A)} \}$.
\end{itemize}

     \noindent \textbf{Premise $1$:} 

     \noindent $\gamma(S_1, \; \mathit{Unstack(A,B)}) \rightarrow S_2$
     
     \smallskip
     \noindent In the current state $\{ \mathit{Ontable(C)}, \mathit{On(B,C)}, \mathit{On(A,B)}, \mathit{Clear(A)} \}$, we can execute the action $\mathit{Unstack(A,B)}$, that results in the next state $\{ \mathit{Ontable(C)}, \mathit{On(B,C)}, \mathit{Clear(A)}, \mathit{Clear(B)}, \mathit{Ontable(A)} \}$.
     
     \smallskip
     \noindent \textbf{Premise $2$:} 
     
     \noindent $\mathit{HoldGoal}(\mathit{Ontable(A)}, \; S_2)$. 
     
     \smallskip
     \noindent In the next state $\{ \mathit{Ontable(C)}, \mathit{On(B,C)}, \mathit{Clear(A)}, \mathit{Clear(B)},$
     
     \noindent $\mathit{Ontable(A)} \}$, the goal $\mathit{Ontable(A)}$ holds.
     
          \smallskip
     \noindent \textbf{Conclusion:} 
     
     \noindent $\mathit{AchieveGoal(\mathit{Unstack(A,B)}, \mathit{Ontable(A)})}$: 
     
     \smallskip
     \noindent Therefore, the action $\mathit{Unstack(A,B)}$ achieves the goal $\mathit{Ontable(A)}$. 
     
\hfill$\Box$
\end{example}

\begin{definition}{(Plan Summary Argument Scheme $\mathit{Arg}_{\pi}$)}
A plan summary argument $\mathit{Arg}_{\pi}$ explains that a proposed sequence of actions $\pi=\langle a_1,a_2,...,a_n \rangle$ is a solution to the planning problem $P$ because it achieves a set of goals $G$:

\begin{itemize}
    \item \textbf{Premise $1$:} 
    $\gamma(S_1,a_1) \rightarrow S_2$, $\gamma(S_2,a_2) \rightarrow S_3$,...,$\gamma(S_n,a_n) \rightarrow S_{n+1}$.
    In the initial state $S_1 = \bigtriangleup_I$, we can execute the first action $a_1$ in the sequence of actions $\pi$ that results in the next state $S_2$ and execute the next action $a_2$ in the sequence in the state $S_2$ that results in the next state $S_3$ and carry on until the last action $a_n$ in the sequence is executed in the state $S_n$ that results in the goal state $S_{n+1}=\bigtriangleup_G$.
    \item \textbf{Premise $2$:} $\mathit{HoldGoals(G, \bigtriangleup_G)}$. In the goal state $\bigtriangleup_G$, all the goals in the set of goals $G$ hold.
    \item \textbf{Premise $3$:} $\mathit{AchieveGoals(\pi,G)}$. The sequence of actions $\pi$ achieves the set of all goals $G$.      
    \item \textbf{Conclusion:} $\mathit{Solution(\pi, P)}$. Therefore, $\pi$ is a solution to the planning problem $P$.
\end{itemize}
\end{definition}

\begin{example}
\label{example:plan-as}

The plan summary argument $\mathit{Arg}_{\pi}$ for the solution plan given in the Example~\ref{example:blocks-world} is shown as follows. Where:
\begin{itemize}
    \item $S_1 = \bigtriangleup_I = \{ \mathit{Ontable(C), On(B,C), On(A,B), Clear(A)} \}$
    \item $S_2 = \{ \mathit{On(B,C), Clear(A), Clear(B), Ontable(A), Ontable(C)} \}$
    \item $S_3 = \{ \mathit{Clear(A), Clear(B), Clear(C), Ontable(A), Ontable(B),}$
    
    $\mathit{Ontable(C)} \}$
    \item $S_4 = \bigtriangleup_G = \{ \mathit{On(C,A)}, \mathit{Ontable(A)}, \mathit{Ontable(B)}, \mathit{Clear(C)},$
    
    $\mathit{Clear(B)} \}$
    \item $G = \{ \mathit{On(C,A)}, \mathit{Ontable(A)}, \mathit{Ontable(B)}, \mathit{Clear(C)}, \mathit{Clear(B)} \}$
    \item $\pi = \langle \mathit{Unstack(A,B)}, \mathit{Unstack(B,C)}, \mathit{Stack(C,A)} \rangle$
\end{itemize}
    
    \smallskip
     \noindent \textbf{Premise $1$:} 
     



    \noindent $\gamma(S_1, \mathit{Unstack(A,B)}) \rightarrow S_2$
     
     \smallskip

     \noindent In the initial state $\{ \mathit{Ontable(C), On(B,C), On(A,B), Clear(A)} \}$, we can execute the action $\mathit{Unstack(A,B)}$ that results in the next state $\{ \mathit{On(B,C), Clear(A), Clear(B), Ontable(A), Ontable(C)} \}$.
     
    \smallskip
    
    \noindent $\gamma(S_2, \mathit{Unstack(B,C)}) \rightarrow S_3$
     
     \smallskip
     \noindent In the state $\{ \mathit{On(B,C), Clear(A), Clear(B), Ontable(A), Ontable(C)} \}$, we can execute the action $\mathit{Unstack(B,C)}$ that results in the next state $\{ \mathit{Clear(A), Clear(B), Clear(C), Ontable(A), Ontable(B),}$
     
     \noindent $\mathit{Ontable(C)} \}$.
     
     \smallskip
     
    \noindent $\gamma(S_3, \mathit{Stack(C,A)}) \rightarrow S_4$

     \smallskip
     \noindent In the state $\{ \mathit{Clear(A), Clear(B), Clear(C), Ontable(A), Ontable(B),}$
     
     \noindent ${Ontable(C)} \}$, we can execute the action $\mathit{Stack(C,A)}$ that results in the goal state $\{ \mathit{On(C,A)}, \mathit{Ontable(A)}, \mathit{Ontable(B)}, \mathit{Clear(C)},$
     
     \noindent $\mathit{Clear(B)} \}$.
     
     \smallskip
      \noindent \textbf{Premise $2$:} 
      
      \noindent $ \mathit{HoldGoals}(\bigtriangleup_G, \; G)$ 
      
      \smallskip
      \noindent In the goal state $\{ \mathit{On(C,A)}, \mathit{Ontable(A)}, \mathit{Ontable(B)}, \mathit{Clear(C)},$
      
      \noindent $\mathit{Clear(B)} \}$, all the goals in the set of goals $\{ \mathit{On(C,A)}, \mathit{Ontable(A)},$
      
      \noindent $\mathit{Ontable(B)}, \mathit{Clear(C)}, \mathit{Clear(B)} \}$ hold.
      
     \smallskip
      \noindent \textbf{Premise $3$:}      
      
      \noindent $\mathit{AchieveGoals}(\pi, \; G)$ 
      
      \smallskip
      \noindent The sequence of actions $\langle \mathit{Unstack(A,B)}, \mathit{Unstack(B,C)}, \mathit{Stack(C,A)} \rangle$ achieves the set of all goals $\{ \mathit{On(C,A)}, \mathit{Ontable(A)}, \mathit{Ontable(B)},$
      
      \noindent $\mathit{Clear(C)}, \mathit{Clear(B)} \}$.    
      
     \smallskip
      \noindent \textbf{Conclusion:}   
      
      \noindent $\mathit{Solution(\pi, \; P)}$
      
      \smallskip
      \noindent Therefore, $\langle \mathit{Unstack(A,B)}, \mathit{Unstack(B,C)}, \mathit{Stack(C,A)} \rangle$ is a solution to the planning problem $P$.
\hfill$\Box$
\end{example}

Having described the schemes and shown how they are used, we turn to the critical questions (CQs). The four CQs given below describe the ways in which the arguments built using the argument schemes can interact with each other. 
These CQs are associated to (i.e., attack) one or more premises of the arguments constructed using the argument schemes and are in turn answered (i.e., attacked) by the other arguments, which are listed in the description.

\smallskip

\noindent \textbf{CQ1: Is it possible for the plan $\pi$ to be a solution?} This CQ begins the dialogue with the user, and it is the first question that the user asks when presented with a solution plan $\pi$. The argument scheme $\mathit{Arg}_{\pi}$ answers the CQ by constructing the summary argument for the plan $\pi$.

\smallskip

\noindent \textbf{CQ2: Is it possible to execute the action $a$?} This CQ is associated with the following argument schemes: $\mathit{Arg}_{\pi}$, $\mathit{Arg}_{S}$, $\mathit{Arg}_{g}$. The argument scheme $\mathit{Arg}_{a}$ answers the CQ by constructing the explanation argument for the action $a$.


\smallskip

\noindent \textbf{CQ3: Is it possible to have the state $S$?} This CQ is associated with the following argument schemes: $\mathit{Arg}_{\pi}$, $\mathit{Arg}_{a}$, $\mathit{Arg}_{g}$. The argument scheme $\mathit{Arg}_{S}$ answers the CQ by constructing the explanation argument for the state $S$.

\smallskip

\noindent \textbf{CQ4: Is it possible to achieve the goal $g$?} This CQ is associated with the argument scheme $\mathit{Arg}_{\pi}$. The argument scheme $\mathit{Arg}_{g}$ answers the CQ by constructing the explanation argument for the goal $g$.

We organise the arguments and their interactions by mapping them into a Dung abstract argumentation framework~\cite{Dung95onthe} denoted by $\mathit{AAF} = (\mathcal{A}, \mathcal{R})$, where $\mathcal{A}$ is a set of arguments and $\mathcal{R}$ is an attack relation $(\mathcal{R} \subseteq \mathcal{A} \times \mathcal{A})$. $\mathit{Args} \subset \mathcal{A}$ and $\mathit{CQs} \subset \mathcal{A}$, where $\mathit{Args} = \{\mathit{Arg}_{\pi}, \mathit{Arg}_a, \mathit{Arg}_S, \mathit{Arg}_g\}$ and $\mathit{CQs} = \{\mathit{CQ}_1, \mathit{CQ}_2, \mathit{CQ}_3, \mathit{CQ}_4\}$.
Given the way that the plan explanation arguments were constructed, they will be acceptable under the grounded semantics if the plan is valid.
We present the properties of the plan explanation arguments as follows.

\noindent \textbf{Property 5.1.}
For a valid plan $\pi$, the set of arguments $\mathit{Args}$ is complete, in that, if a $\mathit{CQ} \in \mathit{CQs}$ exists, then it will be answered (i.e., attacked) by an $\mathit{Arg} \in \mathit{Args}$.

\begin{proof}
Since, $(\mathit{Arg}_{\pi}, CQ1) \in \mathcal{R}$, $(\mathit{Arg}_a, CQ2) \in \mathcal{R}$, $(\mathit{Arg}_S, CQ3) \in \mathcal{R}$, and $(\mathit{Arg}_g, CQ4) \in \mathcal{R}$, therefore, a unique $\mathit{Arg} \in \mathit{Args}$ exists, that attacks a unique $\mathit{CQ} \in \mathit{CQs}$. Thus, $\mathit{Args}$ is complete. 
\end{proof}
In other words, if the plan is valid, then all the objections that can be put forward regarding the plan and its elements do not hold. In particular:

\noindent \textbf{Property 5.2.}
For a valid plan $\pi$, $\mathit{Arg}_{\pi} \in \mathit{Gr}$ iff $CQ \not\in \mathit{Gr}$ when $(\mathit{CQ}, \mathit{Arg}_{\pi}) \in \mathcal{R}$, $CQ \in CQs$.

\begin{proof}
Follows from Property 5.1. Since any CQ that attacks $\mathit{Arg}_{\pi}$ is in turn attacked by an $\mathit{Arg} \in \mathit{Args}$, therefore, $CQ \not\in \mathit{Gr}$. Thus, $\mathit{Arg}_{\pi} \in \mathit{Gr}$. 
\end{proof}

In a very similar way we can show the following:

\noindent \textbf{Property 5.3.}
For a valid plan $\pi$, $\mathit{Arg}_{\pi} \in \mathit{Gr}$ iff $\forall g \in G \; \mathit{Arg}_g \in \mathit{Gr}$.


\smallskip

\noindent \textbf{Property 5.4.}
For a valid plan $\pi$, $\mathit{Arg}_{\pi} \in \mathit{Gr}$ iff $\forall a \in A \; \mathit{Arg}_a \in \mathit{Gr}$.


\smallskip

\noindent \textbf{Property 5.5.}
For a valid plan $\pi$, $\mathit{Arg}_{\pi} \in \mathit{Gr}$ iff $\forall S_i \in S \; \mathit{Arg}_{S_i} \in \mathit{Gr}$.


\smallskip

\noindent
In other words, all the objections of the user regarding: the set of goals $G$; the actions $A$; and states $S$ (where $\exists S_i \in S$ held at each action step) that are related to the plan $\pi$, do not hold.

\smallskip

\noindent \textbf{Property 5.6.}
For a plan $\pi$, $\mathit{Arg}_{\pi} \in \mathit{Gr}$ iff plan $\pi$ is valid.~\footnote{Follows immediately from properties 5.1--5.5.}

In other words, all the objections regarding the plan $\pi$ and its elements do not hold. Therefore, plan $\pi$ is valid.

\noindent \textbf{Property 5.7.}
For a plan $\pi'$, $\mathit{Arg}_{\pi'} \not\in \mathit{Gr}$ iff plan $\pi'$ is invalid.~\footnote{Follows immediately from properties 5.1--5.5.}


Collectively these properties align the notion of a valid plan with an acceptable plan explanation argument.
If we assume, as we do going forward, that a user is rational in the sense of 1) accepting the arguments in the grounded extension of a Dung framework, and 2) only holding to true those facts in the planning model, then if a plan is valid, that user will accept that the plan explanation argument holds.
It is in that sense that we consider the argument to be a suitable explanation --- it can justify all the plan elements by providing arguments that are grounded in the planning model.

\section{Dialogue System using Argument Schemes and Critical Questions}
\label{sec:dialogue-game}

In this section, we present a system for a formal dialogue between planner and user that allows the user to explore the plan explanation arguments, raising objections and having them answered.
It provides an interactive process that recursively unpacks the explanation arguments and allows the user to assure themselves of their acceptability (and hence the validity of the plan).


The dialogue takes place between two participants: \begin{inparaenum}
\item planner; and \item user
\end{inparaenum}. The communication language consists of the legal dialogue moves by the two participants. The moves that the planner and user can use in a dialogue are defined as follows.
\begin{definition}{(Planner Moves)}
Planner moves consist of the following:
\begin{inparaenum}
\item $\mathit{Arg_{\pi}}$, plan summary argument.
\item $\mathit{Arg_a}$, action argument.
\item $\mathit{Arg_S}$, state argument.
\item $\mathit{Arg_g}$, goal argument.
\end{inparaenum}
\end{definition}

\begin{definition}{(User Moves)}
User moves consist of the following:
\begin{inparaenum}
\item $\mathit{CQ_1}$, is it possible for the plan $\pi$ to be a solution?
\item $\mathit{CQ_2}$, is it possible to execute the action $a$?
\item $\mathit{CQ_3}$, is it possible to have the state $S$?
\item $\mathit{CQ_4}$, is it possible to achieve the goal $g$?
\end{inparaenum}
\end{definition}
A joint commitment store is used for holding the planner and user moves (i.e., arguments) used within a dialogue.
\begin{definition}{(Commitment Store)}
A \textit{commitment store} denoted by $\mathit{CS} \subseteq (\mathit{Args} \cup \mathit{CQs})$ holds all the arguments (i.e., planner moves and user moves) which the planner and user are dialectically committed to. 
 $\mathit{CS(pl)}$ denotes all the arguments of the planner and $\mathit{CS(us)}$ denotes all the arguments of the user.
\end{definition}
To ensure that the dialogue conversation ends, we define the termination conditions for the dialogue as follows.
\begin{definition}{(Termination conditions)}
The dialogue terminates when any one of the following three conditions holds:
\begin{itemize}
    \item $T_1: \mathit{PlannerMove} = null$. When the planner is unable to generate the argument, because one of the premises of the argument is not true.
    \item $T_2: \mathit{UserMove} = null$. When the user has exhaustively asked all CQs regarding the components of the plan.
    \item $T_3: \mathit{UserMove} = none$. When the user does not want to ask any more questions.
\end{itemize}
\end{definition}

Termination of the dialogue conversation results in an outcome.
\begin{definition}{(Dialogue Outcomes)}
There are three possible outcomes of the dialogue:
\begin{itemize}
    \item $O_1 =$``Plan is invalid and explanation is unacceptable".
    \item $O_2 =$ ``Plan is valid and explanation is acceptable".
    \item $O_3 =$ ``Explanation is acceptable".
\end{itemize}
\end{definition}
The user and planner both have to follow rules.
%
\begin{definition}{(User Move Rules)}
\label{def:user-move-rules}
The moves that are allowed for the user to put forward depend on the previous moves of the planner~\footnote{User can select any one of the previous moves put forward by the planner.}. The allowed user moves in response to the planner moves are given below, where the first move does not require any previous planner moves.
\begin{inparaenum}
    \item $\mathit{CQ_1}$~\footnote{We assume that planner has already presented the plan $\pi$ to the user and do not consider this a move.}
    \item $\mathit{Arg_a}$: $\mathit{CQ_3}$
    \item $\mathit{Arg_S}$: $\mathit{CQ_2}$
    \item $\mathit{Arg_g}$: $\mathit{CQ_2, CQ_3}$    
    \item $\mathit{Arg_{\pi}}$: $\mathit{CQ_2, CQ_3, CQ_4}$  
\end{inparaenum}

\end{definition}
%
\begin{definition}{(Planner Move Rules)}
\label{def:planner-move-rules}
The moves that are allowed for the planner to put forward depend on the previous move by the user. The allowed planner moves in response to the user moves are given below.
\begin{inparaenum}
    \item $\mathit{CQ_1}$: $\mathit{Arg_{\pi}}$
    \item $\mathit{CQ_2}$: $\mathit{Arg_a}$
    \item $\mathit{CQ_3}$: $\mathit{Arg_S}$
    \item $\mathit{CQ_4}$: $\mathit{Arg_g}$    
\end{inparaenum}
\end{definition}

The dialogue has to follow certain rules (i.e., a protocol).
\begin{definition}{(Dialogue Rules)}
Following are the rules of the dialogue denoted by $\mathit{DR}$:
\begin{inparaenum}
    \item The first move in the dialogue is made by the user, which is, $\mathit{CQ_1}$.
    \item Both players, i.e., planner and user can put forward a single move at a given step in response to each other.
    \item Once the move is put forward, it is stored in the commitment store $\mathit{CS}$.
    \item The user cannot put forward a move, i.e., argument already present in the commitment store $\mathit{CS}$ for a plan component and the same goes for the planner.~\footnote{Since a previously asked user CQ for a plan component is not allowed, therefore, we assume that the planner will not repeat the same argument.}
    \item Each user move has to follow the user move rules given in Definition~\ref{def:user-move-rules}.
    \item Each planner move has to follow the planner move rules given in Definition~\ref{def:planner-move-rules}.    
    \item The dialogue ends when any one of the termination conditions $T_1$, $T_2$, or $T_3$ holds.
\end{inparaenum}

\end{definition}

The dialogue between the planner and user is then defined.
\begin{definition}{(Dialogue)}
We define a dialogue to be a sequence of moves $\mathcal{D}= [M_0,M_1,...,M_n]$. The dialogue takes place between the two participants, i.e., planner and user. Each dialogue participant must follow the dialogue rules $\mathit{DR}$ for making moves. Each move put forward by both participants is recorded and stored in the commitment store $\mathit{CS}$. The dialogue terminates when any one of the termination conditions $T_1, T_2,$ or $T_3$ holds. Based on the termination condition $T$, the outcome of the dialogue can be:
\begin{itemize}
    \item If $T=T_1$, then outcome of the dialogue is: 
    
    $O_1 =$``Plan is invalid and explanation is unacceptable".
    \item If $T=T_2$, then outcome of the dialogue is:
    
    $O_2 =$ ``Plan is valid and explanation is acceptable".
    \item If $T=T_3$, then outcome of the dialogue is:
    
    $O_3 =$ ``Explanation is acceptable".
\end{itemize}
\end{definition}

In order to describe the dialectical process, we introduce Algorithms 1--4. Algorithm~\ref{alg:dialogue-game} presents an operational description of the dialogue $\mathcal{D}$ in the dialogue system specified above, that allows the planner and user to find moves, i.e., arguments to put forward in a dialogue, based on the counterarguments put forward by the other party. The dialogue starts when the planner presents a plan $\pi$ to the user and the user asks the first $CQ_1$. During the dialogue conversation, the planner is either able to construct an appropriate argument via the argument schemes; or the argument returned is \textit{null}, indicating that it is unable to construct the argument, i.e., $T_1$ holds, which consequently \textit{terminates} the dialogue. Similarly, the user either has the option of asking a critical question regarding the premises of any one argument from the previous arguments put forward by the planner; or the user has no more questions to ask, i.e., $T_3$ holds, which consequently \textit{terminates} the dialogue. Furthermore, the user may exhaustively ask all the critical questions (and the planner successfully answers them), i.e., $T_2$ holds, and thus, the dialogue \textit{terminates}. The dialogue finishes with one of the three possible \textit{outcomes} $O_1$, $O_2$, or $O_3$.

Algorithm~\ref{alg:user-moves}--\ref{alg:select-user-moves} provide an operational description of ways that a rational user may behave. We can think of these as both a mechanism for proving that the dialogue works as desired, and as a mechanism for generating allowable moves in an implementation that walks the user through an exploration of the plan.
In Algorithm~\ref{alg:user-moves}, the user is able to choose a previous planner argument in the dialogue and find all the allowed moves that she can put forward in response to that. The input of Algorithm~\ref{alg:user-moves} is the set of all previous planner moves, i.e., arguments generated via the argument schemes and the output is a set of user moves consisting of arguments, i.e., CQs, from which the user can further choose a single CQ. In Algorithm~\ref{alg:select-user-moves}, the user can select a particular move from the set of user moves or choose not to ask any more questions. Thus, the output of Algorithm~\ref{alg:select-user-moves} is either the chosen CQ or \textit{none}. In Algorithm~\ref{alg:planner-move}, the planner can find the relevant move (i.e., argument generated via the argument schemes) to answer the user CQ. The input of Algorithm~\ref{alg:planner-move} is the previous user move, i.e., CQ and the output is an appropriate planner move corresponding to the user CQ. If the planner is unable to generate the argument via the argument schemes, then we assume the output planner move to be \textit{null}.

\begin{algorithm}[!h]
	\begin{algorithmic}[1]
		\Require $\mathit{Args_P}$, set of previous arguments put forward by planner
		
		\Ensure $\mathit{UserMoves}$, set of all allowed user moves
		\Function{FindUserMoves}{$\mathit{Args_P}$}
		\If{$\mathit{Args_P} \neq \emptyset$}
		\State $\mathit{Arg} \gets$ user chooses $\mathit{Arg}$ s.t. $\mathit{Arg} \in \mathit{Args_P}$
		\If{$\mathit{Arg} = \mathit{Arg_a}$} \Comment if $Arg$ is the action  argument
		\State $\mathit{UserMoves} \gets \{\mathit{CQ_3}\}$
		\ElsIf{$\mathit{Arg} = \mathit{Arg_S}$} \Comment if $Arg$ is the state  argument
		\State $\mathit{UserMoves} \gets \{\mathit{CQ_2}\}$
		\ElsIf{$\mathit{Arg} = \mathit{Arg_g}$} \Comment if $Arg$ is the goal  argument
		\State $\mathit{UserMoves} \gets \{\mathit{CQ_2, CQ_3}\}$
		\ElsIf{$\mathit{Arg} = \mathit{Arg_{\pi}}$} \Comment if $Arg$ is the plan summary argument
		\State $\mathit{UserMoves} \gets \{\mathit{CQ_2, CQ_3, CQ_4}\}$	
		\EndIf
		\Else \Comment if the plan $\pi$ is presented to the user
		\State $\mathit{UserMoves} \gets \{\mathit{CQ_1}\}$	
		\EndIf
		\State \Return $\mathit{UserMoves}$
		\EndFunction
	\end{algorithmic}
	\caption{Find User Moves}
	\label{alg:user-moves}
\end{algorithm}

\begin{algorithm}[!h]
	\begin{algorithmic}[1]
		\Require $\mathit{UserMoves}$, set of all allowed user moves
		\Require $\mathit{CQs_U}$, set of previous arguments put forward by user
		\Ensure $\mathit{UserMove}$, a selected user move
		\Function{SelectUserMove}{$\mathit{UserMoves}, \mathit{CQs_U}$}
		\If{(User wants to question further) $\&\& \; (\mathit{UserMoves} \not\subseteq \mathit{CQs_U})$} 
		\State $\mathit{UserMove} \gets$ user chooses $u \; s.t. \; u \in \mathit{UserMoves} \; \&\& \; u \not\in \mathit{CQs_U}$
		\Else \Comment User does not have any more questions
		\State $\mathit{UserMove} \gets \mathit{none}$
		\EndIf
		\State \Return $\mathit{UserMove}$
		\EndFunction
	\end{algorithmic}
	\caption{Select User Move}
	\label{alg:select-user-moves}
\end{algorithm}

\begin{algorithm}[!h]
	\begin{algorithmic}[1]
		\Require $\mathit{UserMove}$, critical question put forward by user
		\Ensure $\mathit{PlannerMove}$, allowed planner move
		\Function{FindPlannerMoves}{$\mathit{UserMove}$}
		\If{$\mathit{UserMove} = \mathit{CQ_1}$} 
		\State $\mathit{PlannerMove} \gets \mathit{Arg_{\pi}}$
		\ElsIf{$\mathit{UserMove} = \mathit{CQ_2}$} 
		\State $\mathit{PlannerMove} \gets \mathit{Arg_a}$
		\ElsIf{$\mathit{UserMove} = \mathit{CQ_3}$} 
		\State $\mathit{PlannerMove} \gets \mathit{Arg_S}$
		\ElsIf{$\mathit{UserMove} = \mathit{CQ_4}$} 
		\State $\mathit{PlannerMove} \gets \mathit{Arg_g}$	
		\EndIf
		\State \Return $\mathit{PlannerMove}$
		\EndFunction
	\end{algorithmic}
	\caption{Find Planner Move}
	\label{alg:planner-move}
\end{algorithm}

\begin{algorithm}[!h]
	\begin{algorithmic}[1]
	    \Require $CQs$, set of all critical questions for plan $\pi$
		\Ensure $\mathit{outcome}$, outcome of the dialogue
		\Function{Dialogue}{$CQs$}
		\State $\mathit{PlannerMove} \gets null$, $\mathit{UserMoves} \gets \emptyset$
		\State $\mathit{UserMove} \gets null$, $\mathit{CS} \gets \emptyset$
		  \Do
		   \If{$\mathit{CS(us)} \neq \mathit{CQs}$} \Comment User has not exhaustively asked all CQs.
          \State $\mathit{UserMoves} \gets \mathit{FindUserMoves(CS(pl))}$
       \State $\mathit{UserMove} \gets \mathit{SelectUserMove(UserMoves, CS(us))}$
    \If{$\mathit{UserMove} \neq \mathit{none}$}
        \State $\mathit{CS} \gets \mathit{CS} \cup \{ \mathit{UserMove} \}$       
       \State $\mathit{PlannerMove} \gets \mathit{FindPlannerMove(UserMove)}$
       \If{$\mathit{PlannerMove} \neq null$}
       \State $\mathit{CS} \gets \mathit{CS} \cup \{ \mathit{PlannerMove} \}$
       \EndIf
    \EndIf
    \Else \Comment User has exhaustively asked all CQs.
        \State $\mathit{UserMove} \gets \mathit{null}$
     \EndIf
  \doWhile{(($\mathit{UserMove} \neq \mathit{null}$) or ($\mathit{UserMove} \neq \mathit{none}$) or ($\mathit{PlannerMove} \neq \mathit{null}$))} 
    \If{$\mathit{PlannerMove} = \mathit{null}$}
        \State $\mathit{outcome} \gets $ ``Plan is invalid and explanation is unacceptable."
    \ElsIf{$\mathit{UserMove} = \mathit{null}$}
        \State $\mathit{outcome} \gets $ ``Plan is valid and explanation is acceptable."
    \ElsIf{$\mathit{UserMove} = \mathit{none}$}
        \State $\mathit{outcome} \gets $ ``Explanation is acceptable."        
    \EndIf
    \State \Return $\mathit{outcome}$
		\EndFunction
	\end{algorithmic}
	\caption{Dialogue}
	\label{alg:dialogue-game}
\end{algorithm}

\subsection{Properties of the Dialogue}
A dialogue $\mathcal{D}$ generated in the dialogue system has the following properties.
In these results, $\mathit{Args_{\mathcal{D}}}$ denotes all the moves, i.e., arguments used by the planner in a dialogue $\mathcal{D}$, where $\mathit{Args_{\mathcal{D}}} \subseteq \mathit{Args}$. $\mathit{CQs_{\mathcal{D}}}$ denotes all the moves, i.e., CQs used by the user in a dialogue $\mathcal{D}$, where $\mathit{CQs_{\mathcal{D}}} \subseteq CQs$.

The first four properties align the validity of a plan with the successful defence of the explanation argument.

\noindent \textbf{Property 6.1.}
For a given dialogue $\mathcal{D}$, plan $\pi$ is valid and its explanation is acceptable iff the planner has exhaustively answered all CQs regarding the plan and its elements, i.e., $\mathit{CQs_{\mathcal{D}}} = CQs$.~\footnote{Follows from property 5.1.}


\noindent \textbf{Property 6.2.}
For a given dialogue $\mathcal{D}$, plan $\pi$'s explanation is acceptable iff the planner has answered all CQs of the user and the user does have any more CQs to ask.~\footnote{Follows from property 6.1.}


\noindent \textbf{Property 6.3.}
For a given dialogue $\mathcal{D}$, plan $\pi'$ is invalid and its explanation is unacceptable to the user iff the planner is unable to answer, i.e., generate an appropriate argument via the argument schemes for at least one $\mathit{CQ} \in \mathit{CQs}_\mathcal{D}$.~\footnote{Follows from property 5.7.}

\noindent \textbf{Property 6.4.}
A dialogue $\mathcal{D}$ for a valid plan $\pi$ results in an explanation that is acceptable.~\footnote{Follows from property 5.6.}


Next we consider termination of the dialogue.

\noindent \textbf{Property 6.5.}
A dialogue $\mathcal{D}$ for a plan $\pi$ always terminates.

\begin{proof}
The three termination conditions $T_1, T_2,$ and $T_3$ ensure that the dialogue $D$ always terminates. We prove this by considering all three conditions:
\begin{inparaenum}
    \item $T_1$: This arises when the planner is unable to construct an appropriate argument in response to a user question. Whenever this happens the dialogue $\mathcal{D}$ terminates.
    \item $T_2$: This arises when the user has asked all possible CQs and the planner has successfully answered them. Whenever this happens the dialogue $\mathcal{D}$ terminates, and the user is not able to put forward any more questions.
    \item $T_3$: This arises when the user has no further questions to ask. Whenever this happens the dialogue $\mathcal{D}$ terminates.
\end{inparaenum}

For a valid plan $\pi$, since $T_2$ condition ensures that the dialogue terminates when the user has exhaustively asked all CQs regarding the elements of the plan (worst-case), and the dialogue rules $DR$ ensure that the user cannot ask the same CQ (for the same plan element) again, therefore, the dialogue $\mathcal{D}$ for a valid plan $\pi$ always terminates.

For an invalid plan $\pi'$, since $T_1$ condition will always arise before $T_2$ which we have already proved above will always terminate, therefore, the dialogue $\mathcal{D}$ for an invalid plan $\pi'$ always terminates.
\end{proof}
Next we show that there is a sense in which the dialogue is sound and complete with respect to the arguments that are expressed.

\smallskip
\noindent \textbf{Property 6.6.}
A dialogue $\mathcal{D}$ for a valid plan $\pi$ is complete, in that, the planner has an argument for each element of the plan and the user has a CQ for each element of the plan. Furthermore, if a user has a $\mathit{CQ} \in \mathit{CQs_{\mathcal{D}}}$ then the planner has a corresponding argument $\mathit{Arg} \in \mathit{Args_{\mathcal{D}}}$ to respond with.

\begin{proof}
Follows from property 5.6. 
Lines $2$-$15$ of Algorithm~\ref{alg:user-moves} help the user in finding a CQ for each plan element corresponding to the planner argument (or the plan). Similarly,  Lines $2$-$10$ of Algorithm~\ref{alg:planner-move} help the planner in finding an argument for each plan element (or plan summary) corresponding to a user CQ. 
\end{proof}

\noindent \textbf{Property 6.7.}
A dialogue $\mathcal{D}$ for a valid plan $\pi$ is sound, in that, $\forall \mathit{CQ} \in \mathit{CQs_{\mathcal{D}}}$, each user move $\mathit{CQ}$ is correct. Similarly, $\forall \mathit{Arg} \in \mathit{Args_{\mathcal{D}}}$, each planner move $\mathit{Arg}$ is correct.

\begin{proof}
Both participants of the dialogue $\mathcal{D}$, i.e., planner and user must follow the dialogue rules $DR$. This ensure that each of them always picks a move that is legally allowed, i.e., correct at every single step in the dialogue $\mathcal{D}$.
Since, Algorithm~\ref{alg:user-moves} finds the set of all allowed user moves following the dialogue rules $DR$ as given in Lines $2$-$15$, therefore $\forall \mathit{CQ} \in \mathit{CQs_{\mathcal{D}}}$, each user move $\mathit{CQ}$ is correct.
Similarly, Algorithm~\ref{alg:planner-move} finds a planner move following the dialogue rules $DR$ as given in Lines $2$-$10$, therefore $\forall \mathit{Arg} \in \mathit{Args_{\mathcal{D}}}$, each planner move $\mathit{Arg}$ is correct.
\end{proof}

\subsection{Example}
\label{sec:illustrative-example}


The following is a dialogue generated by the dialogue system given the blocks world of Example~\ref{example:blocks-world}.
\begin{enumerate}[label=\arabic*)]
    \item The planner presents a plan $\pi$, i.e., sequence of actions, $\pi = \langle \mathit{Unstack(A,B)}, \mathit{Unstack(B,C)}, \mathit{Stack(C,A)} \rangle$ to the user.
    \item Algorithm~\ref{alg:user-moves} is called by the user to find a set of possible moves, and then, Algorithm~\ref{alg:select-user-moves} is called by the user to select a particular move (or end the dialogue if the user has no questions). Since $\pi$ is not an argument, it is assumed that the planner has presented the plan $\pi$. The user asks,
    
    \begin{center}
        \textbf{CQ1: Is it possible for the plan $\pi$ to be a solution?}
    \end{center} 
    \item Algorithm~\ref{alg:planner-move} is called by the planner to find a suitable move, i.e., $\mathit{Arg}_{\pi}$ which is the plan summary argument. Planner presents the argument $\mathit{Arg}_{\pi}$, which is given in Example~\ref{example:plan-as}.~\footnote{$\mathit{Arg}_{\pi}$ is not repeated here and can be seen in Example~\ref{example:plan-as}.}
    \item Algorithm~\ref{alg:user-moves} is called by the user which returns a set of CQs, $\{ CQ_2, CQ_3, CQ_4 \}$, and after that, Algorithm~\ref{alg:select-user-moves} is called by the user to select one of the CQs or end the dialogue. The user asks,
    \begin{center}
        \textbf{CQ2: Is it possible to execute the action $a = \mathit{UNSTACK(A,B)}$?}
    \end{center}
    \item Algorithm~\ref{alg:planner-move} is called by the planner to find a suitable move, i.e., $\mathit{Arg}_a$, which is the action argument for action 
    
    \noindent $a=\mathit{UNSTACK(A,B)}$, as given in Example~\ref{example:action-as}.~\footnote{$\mathit{Arg}_a$ is not repeated here and can be seen in Example~\ref{example:action-as}.}
    \item Algorithm~\ref{alg:user-moves} is called by the user, where the user chooses the previous planner argument $\mathit{Arg}_a$ to question~\footnote{At this point the user can choose any previous planner argument including the plan summary argument $\mathit{Arg}_{\pi}$ to question further.},  which returns a set of CQs, $\{CQ_3\}$. After that, Algorithm~\ref{alg:select-user-moves} is called by the user to select a CQ or terminate the dialogue. The user decides to terminate the dialogue, and thus, finds the explanation acceptable.
\end{enumerate}

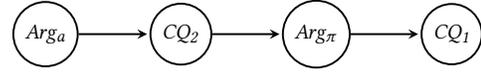
\begin{figure}[t]
	\centering
	\scalebox{0.9}{\begin{tikzpicture}[
	> = stealth, 
	shorten > = 1pt, 
	auto,
	node distance = 2cm, 
	thick 
	]
	
	\tikzstyle{every state}=[
	draw = black,
	thick,
	fill = white,
	minimum size = 4mm
	]
	\node[state] (B)  {$\mathit{Arg_a}$};
	\node[state] (C) [right of=B] {$\mathit{CQ_2}$};
	\node[state] (D) [right of=C] {$\mathit{Arg_{\pi}}$};
	\node[state] (E) [right of=D] {$\mathit{CQ_1}$};
	
	\path[->] (B) edge node {} (C);
	\path[->] (C) edge node {} (D);        
	\path[->] (D) edge node {} (E);
	
	\end{tikzpicture}}
	\caption{Argumentation graph of the example dialogue}
	\label{fig:dialogue_arg_graph}
\end{figure}

Figure~\ref{fig:dialogue_arg_graph} presents the argumentation graph of the above dialogue, where the grounded extension $\mathit{Gr} = \{\mathit{Arg_a}, \mathit{Arg_{\pi}} \}$. The \textit{outcome} of the dialogue is \textbf{``Explanation is acceptable"}, since the user has no further questions to ask and the plan $\pi$ has not yet proven to be invalid. On the other hand, if at any point the planner is unable to construct a suitable argument then the explanation will be considered as unacceptable and this would imply the plan is invalid. Furthermore, if the user has exhaustively asked all CQs and the planner has successfully answered them, then the explanation will be considered acceptable and the plan valid.

\section{Conclusions and Future Work}
\label{sec:conclusions}
We have presented a novel argument scheme-based approach for generating interactive explanations in the domain of AI planning. The novelty of our approach compared to previous research is: 
\begin{inparaenum}[(1)]
\item we have presented novel argument schemes to generate the arguments that directly provide an explanation;
\item we have used the concept of critical questions to allow interaction between the arguments;
\item we have presented a novel dialogue system using the argument schemes and critical questions to provide dialectical interaction between the planner and user; and
\item our approach helps in determining the validity of the plans and whether the explanation is acceptable to a rational user.
\end{inparaenum} 
Note that, our approach to generating explanation arguments is planner independent and therefore, can work on a wide range of input plans in classical planning, and in the future, we intend to extend this to richer planning formalisms such as partial order and temporal planning.



In the future, we aim to develop algorithms based on the argument schemes to automatically extract the arguments from the input planning model. Furthermore, we aim to extend the arguments schemes and CQs to generate contrastive explanations~\cite{ai-Miller19}, and to explore dialogue strategies~\cite{Medellin-GasqueABM13} for finding good moves in a dialogue, for instance, to facilitate the user to find relevant information about the plan with a minimum number of moves. Another avenue of future research is to model user trust~\cite{OgunniyeTO17} during a dialogue conversation and determine how the user trust ratings of the planner explanation arguments should affect the outcome of the dialogue.







\bibliographystyle{ACM-Reference-Format} 
\bibliography{refs}

\end{document}